\documentclass[sigconf]{acmart}

\usepackage{multirow}
\AtBeginDocument{%
  \providecommand\BibTeX{{%
    \normalfont B\kern-0.5em{\scshape i\kern-0.25em b}\kern-0.8em\TeX}}}

\copyrightyear{2023}
\acmYear{2023}
\setcopyright{acmlicensed}
\acmConference[MM '23]{Proceedings of the 31st ACM International Conference on Multimedia}{October 29-November 3, 2023}{Ottawa, ON, Canada}
\acmBooktitle{Proceedings of the 31st ACM International Conference on Multimedia (MM '23), October 29-November 3, 2023, Ottawa, ON, Canada} 
\acmPrice{15.00}
\acmDOI{10.1145/3581783.3611964} 
\acmISBN{979-8-4007-0108-5/23/10}

\acmSubmissionID{1209}

\begin{document}

\title[Enhancing Multi-modal Multi-hop QA via Structured Knowledge and Unified Retrieval-Generation]{Enhancing Multi-modal Multi-hop Question Answering via Structured Knowledge and Unified Retrieval-Generation}



\author{Qian Yang}
\email{qianyang.nlp.cs@gmail.com}
\affiliation{%
\institution{Harbin Institute of Technology}
  \city{Shenzhen}
  \country{China}
}

\author{Qian Chen}
\email{lukechan1231@gmail.com}
\affiliation{%
\institution{Unaffiliated}
  \city{Hangzhou}
  \country{China}
}

\author{Wen Wang}
\email{wwang.969803@gmail.com}
\affiliation{%
\institution{Unaffiliated}
  \city{Hangzhou}
  \country{China}
}

\author{Baotian Hu}
\email{hubaotian@hit.edu.cn}
\authornote{Corresponding author.}
\affiliation{%
\institution{Harbin Institute of Technology}
  \city{Shenzhen}
  \country{China}
}

\author{Min Zhang}
\email{zhangmin2021@hit.edu.cn}
\affiliation{%
\institution{Harbin Institute of Technology}
  \city{Shenzhen}
  \country{China}
}


\begin{abstract}
Multi-modal multi-hop question answering involves answering a question by reasoning over multiple input sources from different modalities. 
Existing methods often retrieve evidences separately and then use a language model to generate an answer based on the retrieved evidences, and thus do not adequately connect candidates and are unable to model the interdependent relations during retrieval.
Moreover, the pipelined approaches of retrieval and generation might result in poor generation performance when retrieval performance is low. 
To address these issues, we propose a Structured Knowledge and Unified Retrieval-Generation (\textbf{SKURG}) approach. 
SKURG employs an Entity-centered Fusion Encoder to align sources from different modalities using shared entities.
It then uses a unified Retrieval-Generation Decoder to integrate intermediate retrieval results for answer generation and also adaptively determine the number of retrieval steps.
Extensive experiments on two representative multi-modal multi-hop QA datasets \emph{MultimodalQA} and \emph{WebQA} demonstrate that SKURG outperforms the state-of-the-art models in both source retrieval and answer generation performance with fewer parameters\footnote{Our code is available at \url{https://github.com/HITsz-TMG/SKURG}}.
\end{abstract}



\maketitle
\section{Introduction}

People build a coherent understanding of the world by actively exploring and seamlessly reasoning over multiple evidences (\emph{multi-hop}) from different modalities (\emph{multi-modal}) which provide complementary information about the environment.
Motivated by this fact, prior works~\cite{talmor2020multimodalqa,chang2022webqa,reddy2022mumuqa} propose the Multi-modal Multi-hop Question Answering (MMQA) task to promote answering \emph{complex} questions by integrating information across different modalities, retrieving the relevant evidence in different modalities, and reasoning across them to provide a correct answer.

\begin{figure}[t]
  \centering
  \includegraphics[width=\linewidth]{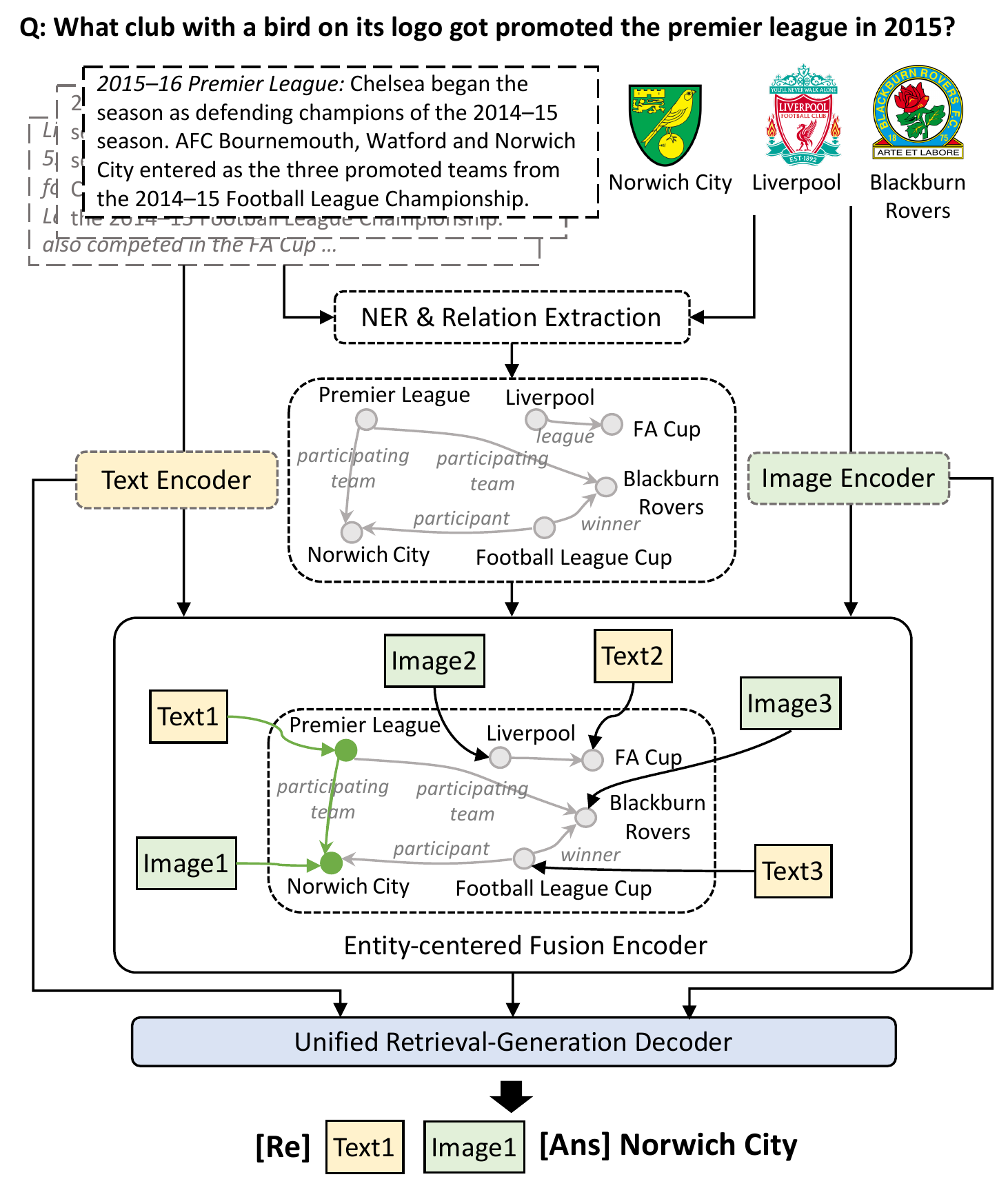}
  \caption{Overview of our proposed SKURG. We use an Entity-centered Fusion Encoder to connect input sources from different modalities and a Unified Retrieval-Generation Decoder to conduct retrieval and answer generation interdependently.}
  \label{fig:intro}
\end{figure}


Existing MMQA systems have two major limitations.
Note that MMQA contains interdependent reasoning steps relying on retrieving accurate evidences~\cite{trivedi2022musique}.
The first limitation is that existing systems are \textbf{insufficient in connecting the input sources in different modalities and modeling the relations between them during retrieval}. 
Some works~\cite{chang2022webqa,oscar} use a classifier on the encoded source for retrieval.
They can only retrieve evidences separately, resulting in the inability to model the interdependent relations between each retrieval hop and inaccuracies in later-hop retrieval. 
Other works~\cite{MuRAG,UnifiedEmbedding} employ a joint encoder for multiple modalities to find the Top-K nearest neighbors as retrieved evidences.
But they still consider each source separately and thus cannot model the relations between evidences during retrieval. 
This can also lead to unbalanced reasoning processes across modalities because the representations of the same modality are usually closer than those from different modalities, even if there are logical connections between the sources of different modalities.
Consequently, the reasoning process may be biased towards certain modalities instead of cross-modalities and degrades the overall performance.
The second limitation is \textbf{error propagation and incapability of adaptively determining the number of retrieval steps}.
Most existing MMQA models
use a retriever to retrieve relevant evidences and a separate reader to generate the answer based on the retrieved evidences. 
The overall QA performance could degrade significantly due to poor retriever performance.
Additionally, due to the lack of reasoning information returned by the reader, the retriever is unable to adaptively determine when to stop retrieval. 

Recently, some methods have been proposed to improve connecting sources~\cite{min2019knowledge,oguz-etal-2022-unik,yu2022kg,huang2022mixed} or unifying retrieval and generation~\cite{qi2021answering,yavuz2022modeling} for \emph{text-only} multi-hop QA. However,  little effort has been made for MMQA. 
To address these two limitations of MMQA, we propose a Structured Knowledge and Unified Retrieval-Generation model (\textbf{SKURG}) for MMQA, which mainly consists of the \textbf{Entity-centered Fusion Encoder (EF-Enc)} and the \textbf{Unified Retrieval-Generation Decoder (RG-Dec)}.
The overview of SKURG is illustrated in Figure~\ref{fig:intro}.
EF-Enc addresses the first limitation of existing MMQA systems by aligning and integrating unstructured multi-modal inputs via structured knowledge representations.
Specifically, EF-Enc uses named entity recognition (NER) and relation extraction on the input sources to construct a knowledge graph. Then it aligns and fuses the entities with the input sources to learn a shared semantic space before retrieval.
While previous methods rely on external knowledge graphs for \emph{text-only} QA, EF-Enc constructs a graph from input sources, which results in a more relevant graph that incorporates \emph{multi-modal} information.
RG-Dec addresses the second limitation by unifying source retrieval and answer generation into a \emph{single} process using a language model. 
It uses a pointer mechanism to retrieve evidences from encoded sources and fused entity representations and then generates the answer based on the evidences. 
RG-Dec tightly integrates the retrieval and reasoning stages and can adapt to \emph{arbitrary} retrieval hops, unlike previous methods which rely on pre-defined numbers of evidences~\cite{yu2022kg}.  
RG-Dec unifies evidence retrieval and answer generation via a language model, hence is broadly similar to PATHFID~\cite{yavuz2022modeling}.
However, RG-Dec is specifically designed for multi-modal sources and employs a pointer mechanism for evidence retrieval rather than generating whole supporting passages as in PATHFID.

The contributions of our work are three-fold: 
    \textbf{\textit{(1)}} 
    We propose an \textbf{Entity-centered Fusion Encoder (EF-Enc)} to integrate \textbf{multi-modal} knowledge into pre-trained generation models, which maps multi-modal sources to a unified semantic space and alleviates modality bias in multi-modal retrieval. 
    Importantly, EF-Enc is tailored to address the unique challenges posed by Multi-modal QA for modeling relations between multi-modal sources.
    To the best of our knowledge, this is the first work to align and fuse multi-modal sources through structured knowledge representations for MMQA. 
    \textbf{\textit{(2)}} 
    We propose a \textbf{Unified Retrieve-Generation Decoder (RG-Dec)} to effectively integrate intermediate retrieval results for answer generation and adaptively determine the number of retrieval steps. 
    RG-Dec further exploits the fused knowledge representations from the Entity-centered Fusion Encoder.
    RG-Dec achieves efficient multi-modal source retrieval and answer generation for Multi-modal Multi-hop QA.
    To the best of our knowledge, this is the first work to unify multi-modal source retrieval and answer generation for MMQA.
    \textbf{\textit{(3)}} 
    Extensive experiments show that \textbf{SKURG significantly outperforms state-of-the-art (SOTA) systems on two representative MMQA datasets on the accuracy of both retrieval and question answering}. 
    Analysis shows that SKURG significantly improves accuracy of multi-modal QA while bringing smaller gains to or slightly degrading accuracy of single-modal QA.
    We achieved the second place on WebQA leaderboard\footnote{As of August 6, 2023, for submitting the camera-ready paper, SKURG achieves the second place on the WebQA leaderboard, with the same retrieval F$_1$ score and only 0.02 absolute lower on the overall QA score compared to the Top performer.}.
\section{Related Works}
\label{sec:relatedworks}

\textbf{Multi-modal Multi-hop Question Answering}
Several MMQA datasets have been proposed recently.
\citet{talmor2020multimodalqa} introduces the MultimodalQA dataset, which includes texts, tables, and images as inputs and constructs single-modality and multi-modality questions.
\citet{chang2022webqa} constructs the WebQA dataset that includes texts and images as sources and requires reasoning over a \emph{single} modality for each question.
Most MMQA methods encode sources separately and use a classification layer~\cite{chang2022webqa} or maximum inner product search~\cite{MuRAG,li-etal-2022-mmcoqa} to select evidences. 
The retrieved evidences are fed to a decoder for answer generation. 
Other methods~\cite{talmor2020multimodalqa,reddy2022mumuqa} decompose the question and use sub-models to conduct retrieval in a single modality for each question part. 
However, it is non-trivial to reliably judge whether a question requires information from single or multiple modalities to answer based only on the question~\cite{qi2021answering}.
Some works~\cite{yoran-etal-2022-turning,xie-etal-2022-unifiedskg} pretrain models on structured knowledge to improve reasoning abilities over tables and texts, but cannot handle multi-modal sources that include images.
These approaches are also unable to jointly integrate information from different modalities.

\noindent
\textbf{Structured Knowledge for Question Answering}
Some prior works~\cite{khot-etal-2017-answering,Quest,reddy2022mumuqa} use information extraction methods to construct knowledge graphs and improve QA performance despite noisy knowledge graphs, but do not include an evidence retrieval stage.
Several works attempt knowledge-guided retrieval for text-only open-domain QA.
\citet{min2019knowledge} uses Wikidata to construct a graph for input passages and extracts the answer from the passage graph.
Similarly, \citet{asai2019learning} constructs a graph using Wikipedia hyperlinks and utilizes a retriever-reader method to score reasoning paths and generate the answer. 
\citet{lv2020commonsense} proposes a graph-based representation and inference module for evidence re-ordering and question answering using evidence from structured knowledge bases and Wikipedia texts. 
\citet{yu2022kg} uses an external knowledge graph and a graph neural network to re-rank retrieved passages and select Top-K evidences for answer generation. 
\citet{oguz-etal-2022-unik} uses an external knowledge graph to linearize triplets into sentences and generates answers based on encoded triplets and texts.

\noindent
\textbf{Unified Retrieval and Answer Generation}
\citet{qi2021answering} uses a unified encoder to retrieve, answer, and re-rank iteratively. It has to repeat this iterative process for each new piece of retrieved evidence and thus is time-consuming.
\citet{yavuz2022modeling} generates the supporting passages and facts and eventually the factoid answer.

\section{Methodology}
Given a question $Q$ and input sources $S=(S_1, ..., S_n)$, where $n$ denotes the number of input sources and each source can be text with its title or image with its caption. MMQA aims to retrieve evidences from $S$ and generate the answer $A$ 
based on the
evidences.


\begin{figure*}[ht]
  \centering
  \includegraphics[width=\linewidth]{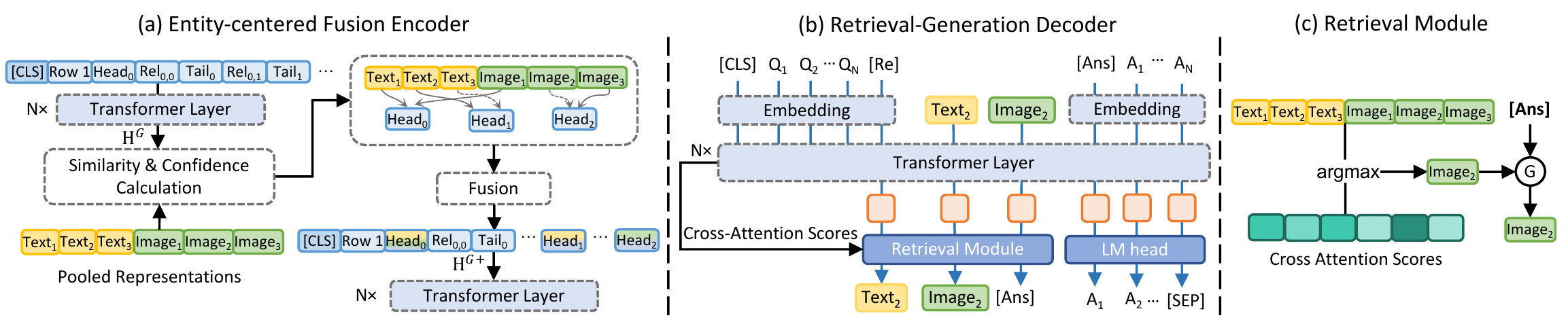}
  \caption{Main modules of SKURG. (a) The Entity-centered Fusion Encoder (EF-Enc) fuses representations of head entities with aligned sources. (b) The Retrieval-Generation Decoder (RG-Dec) integrates intermediate retrieval results for answer generation. (c) The Retrieval Module in (b), which adaptively determines the number of retrieval steps.}
  \label{fig:modules}
\end{figure*}

\subsection{Basic Encoders}
\label{subsec:basic-encoders}
As shown in Figure~\ref{fig:intro}, we use bidirectional Transformer-based~\cite{attnisall} pre-trained models as the text encoder and the image encoder to separately encode input sources.
We prepend the [CLS] token and the question $Q$ to each text source and to each image caption.
After being tokenized into subword tokens~\cite{sennrich2016neural}, 
the text encoder takes the tokenized $i$-th enriched text $T_i=({\rm [CLS]}, q_1, ..., q_{\lvert Q\rvert}, w_{i,1}, ...,$ $ w_{i,m})$ as the input, where $m$ denotes the length of the $i$-th text source.
It then outputs its representation $\textbf{H}_i^T=(\textbf{h}_{i,\rm{[CLS]}}^T, \textbf{h}_{q_1}^T, ...,$ $ \textbf{h}_{q_{\lvert Q\rvert}}^T, \textbf{h}_{w_{i,1}}^T , ..., \textbf{h}_{w_{i,m}}^T)$.
The image encoder takes the $i$-th image $I_i$ and its enriched caption $C_i=({\rm [CLS]}, q_1, ..., q_{\lvert Q\rvert}, c_{i,1}, ..., c_{i,n})$ as the input, where $n$ denotes the length of its original caption.
The image encoder first utilizes ResNet modules~\cite{he2016deep} to convolve the image $I_i$ into $\lvert I_i\rvert = 16\times16 $ patches and extract the corresponding features.
The enriched caption is tokenized into subword tokens similarly and embedded into features. 
The image features with caption embeddings are concatenated and fed to bi-directional Transformer layers.
The output of the image encoder is denoted as $\textbf{H}_i^I=( \textbf{h}_{i,1}^I, ..., \textbf{h}_{i,\lvert I_i\rvert}^I, \textbf{h}_{i,\rm{[CLS]}}^I, \textbf{h}_{q_1}^I, ..., $ $ \textbf{h}_{q_{\lvert Q\rvert}}^I, \textbf{h}_{c_{i,1}}^I, ..., \textbf{h}_{c_{i,n}}^I)$.
We use $\textbf{h}_{i,\rm{[CLS]}}^T$ of each $T_i$ and $\textbf{h}_{i,1}^I$ of each $I_i$ as the pooled representation $\textbf{h}_{i,{\rm{pool}}}$ for each source.

\subsection{Entity-centered Fusion Encoder}
\label{subsec:ef-enc}
The text encoder and image encoder (Section~\ref{subsec:basic-encoders}) only encode sources separately and cannot build connections between them.
We construct a knowledge graph based on the sources $S$ and propose an \textbf{Entity-centered Fusion Encoder} (\textbf{EF-Enc}) to align and fuse sources via structured knowledge, as shown in Figure~\ref{fig:modules}(a).

\noindent \textbf{Knowledge Graph Construction} We first extract entities in the input texts and image captions using a pre-trained named entity recognition (NER) model.
Then we utilize a pre-trained relation extraction model to extract the relations between entities in each source.
Since the input text usually contains many entities, extracting relations for all entities may produce too much noise and obfuscate the key information.
Hence we design the following strategy for a good trade-off between computation cost and overall performance. 
We treat the entities in the text title as the head entities and the entities in the text main body as the tail entities. 
In contrast, each entity in the image caption are treated as both a head entity and a tail entity.
We then extract the relation of each <head entity, tail entity>.
The relation between the $i$-th head entity $e_{i}^h$ and the $j$-th tail entity $e_{j}^t$ is denoted as $r_{i,j}$.
Based on all entity extraction and relation extraction results on input sources $S$, we construct one knowledge graph for all input sources in $S$, as shown in Figure~\ref{fig:intro}, which connects the multi-modal sources via the entities.
Then we follow the baseline approach from MultimodalQA~\cite{talmor2020multimodalqa} and linearize the knowledge graph by concatenating entity pairs and their relations into a sequence. 
We treat each head entity, along with its tail entities and relations, as a row and use ``Row $n$:'' as a delimiter.
For example, this operation converts the graph in Figure~\ref{fig:intro} into the following text: ``[CLS] Row 1: \textit{Premier League: participating team, Norwich City;  participating team, Blackburn Rovers.} Row 2: \textit{Liverpool:} ... '' (Figure~\ref{fig:modules} (a) upper-left). 

\noindent \textbf{Similarity \& Confidence Calculation} We feed the linearized knowledge graph to an embedding layer and bi-directional Transformer layers to obtain the encoded representation, denoted as $\textbf{H}^G=(\textbf{h}_{\rm{[CLS]}}^G, \tilde{\textbf{h}}_{e_{0}^h},  \tilde{\textbf{h}}_{r_{0,0}}, \tilde{\textbf{h}}_{e_{0}^t}, \cdots)$, where $\tilde{\textbf{h}}_{e_{k}^{\{h,t\}}}\in \mathbb{R}^{D}$ denotes the average-pooled representation of the tokens in $e_{k}^{\{h,t\}}$.
To build alignment between sources and head entities,
we compute the inner product between the pooled representations of each source $S_i$ and each head entity $e_k^h$ as the similarity score $s_{i,k}$:
\begin{equation}
\label{eq:sim_score}
    s_{i,k} = (\textbf{h}_{i,\rm{pool}}\textbf{W}^{s})\cdot \tilde{\textbf{h}}_{e_{k}^h}
\end{equation}
$\textbf{h}_{i,{\rm{pool}}}\in \mathbb{R}^{D}$ denotes the pooled representation of the $i$-th input source $S_i$ (Section~\ref{subsec:basic-encoders}), $\textbf{W}^s \in \mathbb{R}^{D\times D}$ is a learnable projection matrix.

Note that $s_{i,k}$ only represents the \emph{relative} similarity of $S_i$ to $e_{k}^h$ among all head entities; hence even the highest $s_{i,k}$ cannot guarantee that $S_i$ is indeed relevant to $e_{k}^h$.
Thus we compute a confidence score to represent the probability that a source should be aligned with the knowledge graph.
Specifically, we concatenate $\textbf{h}_{i,{\rm{pool}}}$ and $\textbf{h}_{\rm{[CLS]}}^G$ and compute the confidence of whether the $i$-th source is related to any head entity in the knowledge graph as follows:
\begin{equation}
\label{eq:threshold}
     {p}_{i} =\sigma([\textbf{h}_{i,{\rm{pool}}} ; \textbf{h}_{\rm{[CLS]}}^G]\textbf{W}^p)
\end{equation}
where $\sigma(\cdot)$ is a sigmoid function, $\textbf{W}^{p}\in\mathbb{R}^{2D\times1}$ is a learnable projection matrix.

Then we form the set $\mathbb{A}_k$ by gathering the sources aligned with $e_{k}^h$.
Note that $S_i$ is added to $\mathbb{A}_k$
if the similarity score between $S_i$ and $e_{k}^h$ is higher than the similarity scores between $S_i$ and other head entities, and also the confidence score of $S_i$ (i.e., $p_i$) is greater than the threshold (a hyperparameter).

\noindent \textbf{Fusion Layer} We fuse the representation of each head entity $e_k^h$ with the pooled representations of all its aligned sources:
\begin{equation}
\label{eq:fusion}
     \tilde{\textbf{h}}_{e_{k}^h}^{\prime} =\tilde{\textbf{h}}_{e_{k}^h}+ \sum_{S_i \in \mathbb{A}_k}{\textbf{h}_{i,{\rm{pool}}}}
\end{equation}

With these operations,
EF-Enc aligns and fuses multi-modal sources through structured knowledge and maps different modalities into a unified semantic space.
For example, given two sources $S_i$ and $S_j$, if a head entity $e_{k}^h$ is more related to $S_i$ and $S_j$ than other head entities (based on similarity score Eq.~\ref{eq:sim_score}) and confidence scores of each $S_i$ and $S_j$ related to the knowledge graph (Eq.~\ref{eq:threshold}) are above the threshold, both $S_i$ and $S_j$ are aligned to $e_{k}^h$ and added to the set $\mathbb{A}_k$.
Their representations are fused into the representation of $e_{k}^h$  (Eq.~\ref{eq:fusion}). Then we update $\textbf{H}^G$ to $\textbf{H}^{G+}$ by replacing $\tilde{\textbf{h}}_{e_{k}^h}$ with $\tilde{\textbf{h}}_{e_{k}^h}^{\prime}$. Finally, we feed $\textbf{H}^{G+}$ to Transformer layers to obtain the encoded representation $\textbf{H}^{G*}$.

\subsection{Retrieval-Generation Decoder}
\label{subsec:rg-dec}
The retrieval-generation decoder (RG-Dec) is illustrated in Figure~\ref{fig:modules}(b).
We adopt a Transformer-based language model as the backbone of RG-Dec.
We feed the question $Q$ to RG-Dec as the prefix information to read the question.
During this \textbf{reading} stage, RG-Dec performs cross-attention over the encoded representations of all sources (Section~\ref{subsec:basic-encoders}) and the \textbf{fused knowledge graph} (Section~\ref{subsec:ef-enc}),  denoted as $\textbf{H}^E=\left(\textbf{H}_1, \cdots, \textbf{H}_n, \textbf{H}^{G*}\right)$.
This helps RG-Dec have a comprehensive understanding of all input sources and consider them holistically to effectively retrieve the relevant evidences in the next stage.
After this stage, RG-Dec begins retrieving evidences from the input sources at time step $\lvert Q\rvert$, i.e., the length of the question. 
A Retrieval Module conducts retrieval and decides when to stop retrieval.
Different from the reading stage, during the \textbf{retrieval} stage, RG-Dec only performs cross-attention over the pooled representations of sources $\textbf{H}_{\rm pool}^E=\left(\textbf{h}_{1,{\rm pool}}, \cdots, \textbf{h}_{n,{\rm pool}}\right)$ (Section~\ref{subsec:basic-encoders}).
This helps RG-Dec better compare each source.
At time step $t$, we choose the source with the highest cross-attention weights as the candidate evidence.
Then we utilize the outputs of the last decoder layer $\textbf{h}_t$ at time step $t$ to decide whether to continue retrieval or to begin answer generation.
Specifically, we calculate a gate score based on  $\textbf{h}_t$:
\begin{equation}
    g_t = \sigma (\textbf{h}_t \textbf{W}^g)
\end{equation}
where $ \textbf{W}^g \in \mathbb{R}^{D\times1}$ denotes a learnable projection matrix. 

If $g_t$ exceeds the threshold, RG-Dec continues with retrieval and uses the pooled representation of the candidate evidence as input for the next time step.
Otherwise, it outputs the token [ANS] and begins answer generation at the next time step.
During the \textbf{answer generation} stage, RG-Dec performs cross-attention over the encoded representations of the retrieved evidences and \textbf{fused knowledge graph}.
For example, if the retrieval module retrieves $S_i$ and $S_j$ as evidences, encoded representations visible to answer generation are $\textbf{H}^E_R=(\textbf{H}_i, \textbf{H}_j, \textbf{H}^{G*})$.
A language model head is used to obtain the generation probability distribution $P_t(a_i \mid \textbf{H}^E_R,a_{<t})$.
Through these three stages, RG-Dec leverages the fused information from EF-Enc to capture interconnections between multimodal sources, enabling more effective retrieval and answer generation.

\subsection{Training}
We train the model with a joint loss.
For EF-Enc, we assume that a source $S_i$ is aligned with a head entity $e_{i^+}^h$ if it contains that entity.
If a source contains multiple head entities, we randomly choose one of the entities to be aligned.
Thus building alignment between sources and entities is converted to a classification problem and we utilize cross-entropy loss for it:
\begin{equation}
    \mathcal{L}_{a}=-\frac{1}{n} \sum_{i=1}^{n} {\log \frac{\exp(s_{i,i^+})}{\sum_{j=1}^{N}\exp(s_{i,j})}}
\end{equation}
where $n$ is the number of sources and $N$ is the number of head entities. 

We also train the confidence estimation of connecting the source to the knowledge graph with the following loss:
\begin{equation}
    \mathcal{L}_{c} \!=\! -\frac{1}{n}\! \sum_{i=1}^{n} {\!(\hat{p}_i \log p_i\!+\!(1\!-\!\hat{p}_i)\log(1\!-\!p_i))}
\end{equation}
where $\hat{p}_i\in \{0,1\}$ indicates whether $S_i$ contains any head entity (\textit{i.e.} 1) or not (\textit{i.e.} 0).

For the retrieval stage in RG-Dec, the decoder begins retrieval at time step $\lvert Q\rvert$, i.e., the length of the question.
We sum the cross-attention scores of all Transformer layers and utilize cross-entropy loss to enforce retrieval in the retrieval steps:
\begin{equation}
    \mathcal{L}_{r}=-\frac{1}{M} \sum_{t=\lvert Q\rvert}^{\lvert Q\rvert+M} {\log \frac{\exp(\alpha_{t, {t^+}})}{\sum_{i=1}^{n}\exp(\alpha_{t,i})}}
\end{equation}
where $\alpha_{t,i}$ denotes the cross-attention scores of the $i$-th source at time step $t$, $\alpha_{t, {t^+}}$ denotes the cross-attention score of the target source at time step $t$, $M$ is the number of retrieval steps.

We also adopt a binary loss to learn when to stop retrieval: 
\begin{equation}
  \begin{split}
    \mathcal{L}_{s}\!=\!-\frac{1}{M}\! \sum_ {\resizebox{.08\hsize}{!}{$t=\lvert Q\rvert$}}^ {\resizebox{.08\hsize}{!}{$\lvert Q\rvert+M$}}\! {(\hat{g}_t\! \log g_t \!+\!(1\!-\!\hat{g}_t\!)\!\log(1\!-\!g_t\!))}
\end{split}
\end{equation}
where $\hat{g}_t\in \{0,1\}$ indicates whether to output the retrieved evidence or end retrieval and output [ANS].

We minimize the negative log-likelihood for the answer generation stage:
\begin{equation}
       \mathcal{L}_{g}\!=\!\sum\nolimits_{t=\lvert Q\rvert+M}^{\lvert Q\rvert+M+\lvert A\rvert}\!-\!\log P_t(a_i \mid \textbf{H}^E_R,a_{<t})
\end{equation}
\noindent where $\lvert A\rvert$ denotes the length of the answer. 

The overall loss function is as follows:
\begin{equation}
    \mathcal{L}= \mathcal{L}_{a}+\mathcal{L}_{c}+\mathcal{L}_{r}+\mathcal{L}_{s}+\mathcal{L}_{g}
\end{equation}

\section{Experiments}
\subsection{Datasets}
\label{subsec:datasets}
We conduct experiments on two representative and most commonly used MMQA datasets: \emph{MultimodalQA} and \emph{WebQA}.
\textbf{MultimodalQA}~\cite{talmor2020multimodalqa} contains multi-modal QA pairs over tables, texts, and images.
There are 16 question types in this dataset, with 13 among them requiring cross-modal retrieval and reasoning~\cite{talmor2020multimodalqa}.
The dataset includes $24$K/$2.4$K QA pairs for training and validation. Since the test set labels are not released, we report SKURG results on the validation set. 
The answers in MultimodalQA are spans or short phrases and the evaluation metrics used are \textbf{Exact Match (EM)} and average \textbf{F}$_1$ as described in~\cite{dua-etal-2019-drop}.
\textbf{WebQA}~\cite{chang2022webqa} contains $34$K/$5$K/$7.5$K QA pairs for training, validation, and testing, where $44\%$ of image-based queries and $99\%$ of text-based questions require two or more input sources. 
The input sources are multi-modal, but all questions require only single-modal retrieval and reasoning.
The WebQA answers are free-form sentences.
Evaluation metrics are \textbf{source retrieval F$_1$} and \textbf{QA} for assessing answer generation quality, which is a combination of the BARTScore~\cite{yuan2021bartscore} based \textbf{QA-FL} and the keyword accuracy score \textbf{QA-Acc}. 
QA-FL measures the fluency (grammaticality and semantic relevance)
between a generated answer and reference
while QA-Acc measures the overlap of key entities between the output answer and reference. 
The \textbf{QA} score is the corpus-level average of the sample-level product of QA-FL and QA-Acc.
The \textbf{QA} score is considered the most important metric for WebQA evaluation, among the various QA metrics.

\subsection{Baselines}
We compare SKURG with SOTA models on MultimodalQA\footnote{We do not compare with models from ~\citet{yoran-etal-2022-turning} on MultimodalQA because they are all \emph{text-only} models and cannot handle images.} and WebQA. 
\textbf{AutoRoute}~\cite{talmor2020multimodalqa} converts multi-modal QA into single-modal QA by using a question-type classifier to identify the modality that might contain the final answer.
It directs the question and input sources to the corresponding QA module (textQ, tableQ, or imageQ) and extracts answer spans using different sub-models. AutoRoute uses RoBERTa-large~\cite{liu2019roberta} for question-type classification and textQ and tableQ, and VILBERT-MT~\cite{VILBERT-MT} for imageQ with image features extracted by Faster R-CNN~\cite{ren2015faster}. 
\textbf{ImplicitDec}~\cite{talmor2020multimodalqa} also uses different sub-models for different modalities. It employs a question-type classifier to identify the relevant modalities, their order, and the logical operations to be applied. During each hop, the corresponding sub-models are used to extract the answer based on the question, inputs of the current modality, and any previously generated answers. Its backbone models are the same as those used in AutoRoute.
\textbf{VLP} and \textbf{VLP + VinVL}~\cite{chang2022webqa} are transformer-based encoder-decoder models with VinVL~\cite{zhang2021vinvl} as the current SOTA method for image feature extraction. They concatenate a source $S_i$ with the question and estimate its selection probability using a classifier. The selected sources and the question are concatenated and fed into the model for answer generation with beam search size 5.
\textbf{MuRAG}~\cite{MuRAG} is pre-trained on a mixture of large-scale image-text and text-only corpora. It uses a query $q$ of any modality to retrieve Top-K nearest neighbors from memory of image-text pairs. 
Retrieved results are combined with $q$ and fed to an encoder-decoder for answer generation. During fine-tuning, it uses question as the query $q$ and Top-4 retrieved sources. Beam search size is 2.  MuRAG uses ViT-large~\cite{vit} to encode images and T5-base~\cite{t5} to encode texts and generate the answer. Note that MuRAG is only evaluated on the text and image subsets of MultimodalQA, filtering out the table modality. 

\subsection{Implementation Details}
We adopt OFA-base encoder\footnote{\url{https://huggingface.co/OFA-Sys/ofa-base}}~\cite{ofa} as the image encoder, which yields SOTA performance on many vision-language tasks while attaining comparable performance on uni-modal tasks.  
We adopt BART-base\footnote{\url{https://huggingface.co/facebook/bart-base}}~\cite{bart} as the text and table encoder and initialize EF-Enc and RG-Dec with BART-base encoder and decoder, respectively as
BART is competitive for text encoding and decoding.
Following AutoRoute and ImplicitDec~\cite{talmor2020multimodalqa}, we pre-train OFA-base and BART-base on SQuAD2.0~\cite{SQuAD} after removing questions without answers.
We adopt ELMo-based NER~\cite{Peters2017SemisupervisedST} for entity extraction and OpenNRE~\cite{opennre} for relation extraction\footnote{Our KG construction introduces little noise with high-accuracy ELMo-based NER~\cite{Peters2017SemisupervisedST} (\textbf{91.9 F$_1$} on CoNLL$_{2003}$ NER task) and BERT$_{base}$-based OpenNRE~\cite{opennre} (\textbf{84.6\% accuracy} on WiKi80 validation set). KG construction is also efficient. ELMo-based NER takes 30s to extract 29,089 entities from 10,000 sources, and BERT$_{base}$-based openNRE takes 350s to extract relations for 39,052 entity pairs from 10,000 sources. These inference times are based on batch\_size=1 on one NVIDIA Quadro RTX 8000.}.
For MultimodalQA, we align the texts and images corresponding to the same table cell
and augment the given table with the linearized extracted knowledge graph. We place the fusion layer between the 4th and 5th layers of BART-base, optimized on MultimodalQA validation set (Table~\ref{tab:multimodalqa_index} in Appendix~\ref{sec:appendix_ablation_fusion_position}).
For training, we use Adam optimizer~\cite{adam} with an initial learning rate of $10^{-5}$ and linear decay for the learning rate.
We set the batch size to 2 and the gradient accumulation steps to 4.
We use 2 NVIDIA A100 40GB for training.
We train 25 epochs for MultimodalQA and 10 epochs for WebQA\footnote{Our computation cost is much lower than MuRAG~\cite{MuRAG}, which uses a batch size of 64 for 20K steps, approx. 106 epochs for MultimodalQA subset and 37 epochs for WebQA.}, selecting the checkpoint with the highest validation scores. Note that the baselines including SOTA MuRAG also select checkpoint this way for reporting results on MultimodalQA validation set.
The threshold for confidence score and gate score is 0.5. The max length of the linearized KG is 760. Details of the training loss analysis is in Appendix~\ref{sec:appendix_loss_function}.

\subsection{Main Results}

\begin{table}
\caption{MultimodalQA results. We report SKURG results on the \emph{validation set}. We cite the \emph{test set} results of baselines from~\citet{talmor2020multimodalqa} and \emph{validation set} results of ImplicitDec from~\citet{yoran-etal-2022-turning}.
Data analysis and baseline results suggest that results on MultimodaQA validation and test sets are comparable (Footnote 9). Best results are in bold.}
\centering
    \resizebox{\linewidth}{!}{
        \begin{tabular}{lccccccc}
        \toprule
         \multirow[m]{2}{*}{Model}    &\multirow[m]{2}{*}{\#Param.}        &\multicolumn{2}{c}{Multi-modal}  &\multicolumn{2}{c}{Single-modal}   &\multicolumn{2}{c}{All}\\
                              &     & EM        &F$_1$      & EM        &F$_1$     & EM   &F$_1$ \\ \midrule
                              & & &\multicolumn{4}{c}{Test Set}\\
                Q-only~\citeyearpar{talmor2020multimodalqa}   &406M   &16.9 &19.5   &  14.2 &17.0   &15.3  & 18.0  \\
                AutoRoute~\citeyearpar{talmor2020multimodalqa}    &1,310M    &32.0 &38.2   & 48.9 &57.1   & 42.1      & 49.5            \\
               ImplicitDec~\citeyearpar{talmor2020multimodalqa}    &1,310M     &46.5 &51.7      & 51.1 &58.8   & 49.3      & 55.9   \\ \midrule
               & & & \multicolumn{4}{c}{Validation Set} \\
                  ImplicitDec~\citeyearpar{talmor2020multimodalqa}    &1,310M     & -- & --      & -- & --   &  48.8       & 55.5 \\
                SKURG (\textbf{ours})     &447M      & \textbf{52.5}    & \textbf{57.2}  & \textbf{66.1}   & \textbf{69.7}  & \textbf{59.8}      & \textbf{64.0}    \\ \bottomrule
        \end{tabular}
    }
\label{tab:multimodalqa_result}
\end{table}

\begin{table}
\caption{Results on subsets of the MultimodalQA \emph{validation set} for different question types. 
We cite baseline results (group 1) from MuRAG~\cite{MuRAG} (MuRAG did not release code and only reported results on Text and Image, excluding Table and multi-modal questions). Best results are in bold.}
\centering
    \resizebox{\linewidth}{!}{
        \begin{tabular}{lcccccc}
        \toprule
         \multirow[m]{2}{*}{Model}      &\multirow[m]{2}{*}{\#Param.}     &\multicolumn{2}{c}{Text}  &\multicolumn{2}{c}{Image}  &Text-Image\\
                           &        & EM        &F$_1$      & EM        &F$_1$     & EM \\ \midrule
                Q-only~\citeyearpar{talmor2020multimodalqa}   &406M    &  15.4 &18.4 &11.0 &15.6 &13.8  \\
                AutoRoute ~\citeyearpar{talmor2020multimodalqa}  &1,310M     & 49.5 &56.9 &37.8 &37.8 &46.6           \\
                MuRAG ~\citeyearpar{MuRAG}    &527M       & 60.8 &67.5 & \textbf{58.2} &\textbf{58.2} &60.2    \\ \midrule
                SKURG (\textbf{ours})    &447M      & \textbf{66.7}   & \textbf{72.7}      & 56.1   &56.1      & \textbf{64.2}    \\ \bottomrule
        \end{tabular}
    }
\label{tab:multimodalqa_result_sub}
\end{table}

We compare SKURG against all the most relevant methods, including the SOTA models.
To ensure a fair and meaningful comparison, we cite the results reported in prior works based on the following rationale. Firstly, all baselines exhibit strong stability, likely due to the large sizes of the datasets and the structures of baselines\footnote{The standard deviation of MuRAG on WebQA test set is quite small as 0.249.}. 
Secondly, we adopt the same data processing steps as the baselines.

\noindent
\textbf{MultiModalQA} We report Multi-modal/Single-modal/All results
in Table~\ref{tab:multimodalqa_result}\footnote{The baseline results for the MultimodalQA validation set were not reported in the original paper~\cite{talmor2020multimodalqa}, and we found that the code for these models was not reproducible. 
However, based on the description of the MultimodalQA dataset~\cite{talmor2020multimodalqa}, the annotation process and distributions between the validation and test sets are \textbf{similar}. 
We found the validation set results of ImplicitDec in~\citet{yoran-etal-2022-turning} from the same authors of the baseline models~\cite{talmor2020multimodalqa}, which shows that the baseline results on the validation set and test set are \textbf{consistent}.
All these suggest that baseline performance on the test set should be \textbf{comparable} to their performance on the validation set.} and 
subset results for different question types in Table~\ref{tab:multimodalqa_result_sub}, where \textbf{Q-only} denotes using only the question as input for BART-large~\cite{bart}.
SKURG results are means from 3 runs with different random seeds. 
The standard deviations for EM and F$_1$ for \emph{All} for SKURG in Table~\ref{tab:multimodalqa_result} are small, as 0.45 and 0.47, respectively, and the standard deviation for Text-Image EM for SKURG in Table~\ref{tab:multimodalqa_result_sub} is also small as 0.62.
Table~\ref{tab:multimodalqa_result} shows SKURG outperforms ImplicitDec by \textbf{11.0 EM/8.5 F$_1$} absolutely on the full validation set, with much fewer parameters (447M vs. 1310M)
and without using the question type. 
Table~\ref{tab:multimodalqa_result_sub} shows that SKURG outperforms the SOTA MuRAG by \textbf{4.0 on EM} 
on the subset of TextQ and ImageQ with fewer parameters (447M vs. 527M).
However, SKURG obtains lower scores on ImageQ than MuRAG, probably because MuRAG uses ViT-large (307M) for image encoding while SKURG uses OFA-base encoder (90M). Since the multi-modal alignment by SKURG is independent on specific pre-trained models, we expect that using ViT-large in SKURG will 
outperform MuRAG on ImageQ.

\begin{table}
\caption{WebQA \emph{test set} results.  
Best results are in bold.}
\centering
    \resizebox{0.95\linewidth}{!}{
        \begin{tabular}{lccccc}
        \toprule
                Model              & \#Param. & QA-FL$\uparrow$       & QA-Acc$\uparrow$     & \textbf{QA}$\uparrow$  & \textbf{Retr-F}$_1$ $\uparrow$           \\ \midrule
                VLP (Q-only)~\citeyearpar{chang2022webqa}  & 220M   & 34.9 &22.2 &13.4  & --         \\
                
                VLP~\citeyearpar{chang2022webqa}  &220M     & 42.6 &36.7 &22.6     &68.9        \\
                VLP + VinVL~\citeyearpar{chang2022webqa} &220M    & 44.2 &38.9 &24.1    &70.9   \\
                MuRAG~\citeyearpar{MuRAG}  &527M   & \textbf{55.7} &54.6 &36.1 &74.6   \\ \midrule
                SKURG (\textbf{ours})  &357M    & 55.4   & \textbf{57.1}   & \textbf{37.7}    & \textbf{88.2}\\\bottomrule 
        \end{tabular}
    }
\label{webqa_result}
\end{table}

\noindent \textbf{WebQA} Table~\ref{webqa_result} shows the WebQA test set results\footnote{Sometimes QA-FL and QA-Acc are higher but the QA score is lower, which is due to the discrepancy between sample-level multiplication and corpus-level averaging.}, where VLP (Q-only) uses only the question as input for VLP. 
SKURG results are means from 3 runs with different random seeds, and standard deviation for \textbf{QA} is small as 0.25. 
Table~\ref{webqa_result} shows that SKURG outperforms all baselines on both QA and retrieval F$_1$ and outperforms SOTA MuRAG by \textbf{1.6} on the overall QA score and \textbf{13.6} on retrieval F$_1$ with fewer parameters (357M\footnote{Since WebQA does not include the table modality, there is no need for EF-Enc to encode tables, hence \#Params. of SKURG on WebQA is smaller than that on MultimodalQA.} vs. 527M). \textbf{Results on both datasets show that SKURG outperforms all baselines, including SOTA MuRAG, on both retrieval and answer generation}.

\subsection{Ablation Studies}
\label{subsec:ablation_study}

\begin{table*}[h!]
\caption{Ablation study of SKURG on the MultimodalQA \textbf{validation set}. 
EF-Enc, RG-Dec and Enc-Re denote \emph{Entity-centered Fusion Encoder}, \emph{Unified Retrieval-Generation Decoder} and \emph{Encoder Retrieval}. Retr-Pre, Retr-Rec and Retr-F1 denote precision, recall and F$_1$ for retrieval. Best results for each metric are in bold.}
\centering
    \resizebox{\textwidth}{!}{
        \begin{tabular}{c|l|ccccc|ccccc|ccccc}
        \toprule
         \multirow[m]{2}{*}{Row} & \multirow[m]{2}{*}{Model}          &\multicolumn{5}{c|}{Multi-modal}  &\multicolumn{5}{c|}{Single-modal}   &\multicolumn{5}{c}{All}\\
                                   & & EM        & F$_1$  &Retr-Pre  &Retr-Rec &Retr-F$_1$ & EM        & F$_1$    &Retr-Pre  &Retr-Rec &Retr-F$_1$  & EM   &F$_1$  &Retr-Pre  &Retr-Rec &Retr-F$_1$\\ \midrule
                1& SKURG &\textbf{52.5} & \textbf{57.2} &86.1 &\textbf{75.7} &\textbf{80.6}  & 66.1 & 69.7 &94.7 &80.2 &86.7   &\textbf{59.8} & \textbf{64.0}  &89.6   & \textbf{77.7}      &\textbf{83.2} \\
                2& KG w/o EF-Enc   &45.9   & 50.5 &\textbf{86.6} &75.4 & \textbf{80.6} & 66.7 &70.5  &94.2 &80.3 & 86.7  &57.2   &61.3  &\textbf{90.0} &77.5  &\textbf{83.2}   \\
                3& EF-Enc w/o KG  &51.6 & 55.6 &85.0 &75.1 & 79.8  & 65.8 & 70.2 & \textbf{94.8} &\textbf{80.6} &\textbf{87.1}  &59.3 & 63.5  &89.1   & 77.5      &82.9 \\ 
             4& w/o (EF-Enc+KG) &44.1 & 49.2 &86.0 &73.9 & 79.5  & 67.4 & 71.3 &94.1 &79.9 &86.4   &56.7	&61.7  &89.5   & 76.5      &82.5 \\
              5& w/o (RG-Dec+KG) &48.8 & 53.1 &-- &-- &--  & 65.1 & 68.7 &-- &-- &--    &57.6	&61.6  &--   & --   &-- \\
            6& w/o (RG-Dec+EF-Enc+KG) & 43.3 & 47.7 &-- &-- &--  & \textbf{67.8} & \textbf{71.5} &-- &-- &--     &56.5	&60.5  &--   & --   &-- \\\midrule
           7& Enc-Re w/o (EF-Enc+KG) &37.7& 42.3 &78.0 &71.2 & 74.4& 62.2 & 66.0 &75.4 & 79.1 &77.2     &51.0	&55.1  &76.8   & 74.7      &75.7 \\
            \bottomrule
        \end{tabular}
    }

\label{tab:multimodal_ablation}
\end{table*}

\begin{table}[h!]
\caption{Ablation study of SKURG on the WebQA \emph{test set}. Retr-F$_1$ denotes retrieval F$_1$.
EF-Enc, RG-Dec and Enc-Re denote \emph{Entity-centered Fusion Encoder}, \emph{Unified Retrieval-Generation Decoder} and \emph{Encoder Retrieval}. The evaluation metrics are described in Section~\ref{subsec:datasets}. Best results are in bold.}
\centering
    \resizebox{0.95\linewidth}{!}{
        \begin{tabular}{lcccc}
        \toprule
                Model            & QA-FL$\uparrow$       & QA-Acc$\uparrow$     & QA$\uparrow$  & Retr-F$_1\uparrow$           \\ \midrule
                 SKURG        & 55.4   & 57.1  & 37.7    & \textbf{88.2}\\
                w/o (EF-Enc+KG)    & 55.5 	& 57.2 	&37.3     & 88.1   \\
                w/o RG-Dec     & \textbf{55.9} 	&\textbf{57.6}	& \textbf{37.8}  & --   \\
                w/o (RG-Dec+EF-Enc+KG)       & 55.6   & 57.3 & 37.5    & --  \\\midrule
                Enc-Re w/o (EF-Enc+KG)  & 51.9 & 55.1  &35.2    & 76.7
                \\\bottomrule
        \end{tabular}
    }
\label{tab:webqa_ablation}
\end{table}

\begin{table}
\caption{Ablation study of oracle entity alignment (i.e., using gold alignments between head entities and sources) and oracle retrieval (i.e., using gold retrieval results) on the WebQA \emph{validation set}. Best results are in bold.}
\centering
    \resizebox{0.95\linewidth}{!}{
        \begin{tabular}{lcccc}
        \toprule
                Model            & QA-FL $\uparrow$       & QA-Acc $\uparrow$     & QA $\uparrow$  & Retr-F$_1\uparrow$           \\ \midrule
                 SKURG        & 46.8	&64.3	&37.1  & 87.5 \\
            + Oracle Entity Alignment    & 46.9	&64.4	&37.2    & 87.5 \\
                + Oracle Retrieval       & \textbf{47.4}	&\textbf{65.2}	&\textbf{37.8}  &\textbf{100} \\
               \bottomrule
        \end{tabular}
    }
\label{tab:webqa_golden}
\end{table}

\subsubsection{\textbf{Efficacy of EF-Enc and RG-Dec}}
We conduct ablation experiments to validate effectiveness of the proposed Entity-centered Fusion Encoder and Unified Retrieve-Generation Decoder in SKURG.

\noindent
\textbf{MultimodalQA} 
Table~\ref{tab:multimodal_ablation} shows the ablation results of the efficacy of EF-Enc and RG-Dec on MultimodalQA validation set.
To ensure fair and valid comparisons, all ablation experiments follow a consistent strategy of selecting the best checkpoint based on the validation set EM scores and reporting the corresponding scores.
(1) Row1 shows SKURG results which augment the given table with the extracted knowledge graph (KG) 
for entity-source alignment and fusion.
It achieves the best performance on multi-modal QA and Retr-recall.
(2) For Row2, we remove EF-Enc, resulting in no entity-source alignment and fusion (Section~\ref{subsec:ef-enc}); the linearized KG is encoded by BART-base. Compared to SKURG, retrieval and single-modal QA performance are comparable but multi-modal QA performance drops dramatically by \textbf{-6.6 EM/-6.7 F$_1$}, despite same Retr-F$_1$, suggesting the importance of alignment and fusion for multi-modal QA.
(3) Row3 removes the extracted KG and only uses the given tables for entity-source alignment and fusion. Compared to Row1, it degrades slightly on single-modal QA performance by -0.3 EM/+0.5 F$_1$/+0.4 Retr-F$_1$,but significantly degrades multi-modal QA performance by \textbf{-0.9 EM/-1.6 F$_1$/-0.8 Retr-F$_1$}, suggesting the crucial role of the extracted KG for answer generation for multi-modal QA.
(4) Row4 removes EF-Enc and KG, and uses a vanilla BART-encoder to encode the table without fusing aligned source representations. This slightly improves single-modal QA (since alignment and fusion in SKURG still introduce some noise), but significantly degrades multi-modal QA performance by \textbf{-8.4 EM/-8.0 F$_1$}, demonstrating that fusing sources with shared entities helps learn a unified semantic space and improves multi-modal QA. 
Row4 also shows comparable retrieval precision but decreased retrieval recall for multi-modal QA, indicating that building connections between sources helps retrieve \emph{relevant} information.
(5) Row5 removes the retrieval module (RG-Dec) and KG, hence the decoder generates the answer based on \emph{all} encoded sources. 
This results in a notable drop in performance compared to Row3, \textbf{-2.8 EM/-2.5 F$_1$} on multi-modal QA and \textbf{-0.7 EM/-1.5 F$_1$} on single-modal QA, showing the importance of filtering sources via retrieval.
(6) In Row6, removing both EF-Enc and retrieval module drastically degrades multi-modal QA performance (similarly to Row4, Row6 improves single-model QA performance since alignment and fusion introduce some noise). Comparing Row6 against Row5 shows that without connecting multi-modal sources via shared entities, using all encoded sources without conducting retrieval could introduce much noise and hence dramatically degrade answer generation performance on multi-modal QA by \textbf{-5.5 EM/-5.4 F$_1$}. 
Notably, comparing Row6 and Row1 shows the proposed EF-Enc with extracted KG and RG-Dec in SKURG achieve absolute \textbf{+9.2 EM} and \textbf{+9.5 F$_1$} gains on multi-modal QA, proving significant and additive gains of EF-Enc and RG-Dec on multi-modal QA.
(7) Row7 over Row1 shows using encoder-based retrieval, similar to VLP~\cite{chang2022webqa} where a classifier selects sources based on their encoded representations, leads to a significant drop in both QA and retrieval performance on \emph{All} by \textbf{-8.8 EM/-7.5 Retr-F$_1$}, demonstrating that retrieving sources separately harms QA and retrieval performance.

\noindent
\textbf{WebQA} 
Table~\ref{tab:webqa_ablation} shows ablation results of EF-Enc and RG-Dec on WebQA test set (averaged over 3 runs). The standard deviations of the \textbf{QA} score range from 0.22 to 0.40. 
Removing EF-Enc and KG degrades retrieval F$_1$ and QA score (Row2 over Row1), indicating that fusing sources with shared entities improves retrieval and answer generation. 
Removing the retrieval module (i.e., w/o RG-Dec)\footnote{WebQA does not provide given tables as MultimodalQA. Therefore, we cannot solely remove the KG while retaining EF-Enc on WebQA, as EF-Enc relies on the extracted KG to utilize structured knowledge for aligning sources.}
slightly improves the QA score (Row3 over Row1, Row4 over Row2), which is inconsistent with the trend on MultimodalQA.
To further understand this discrepancy, 
we report the performance using gold retrieval results (\emph{oracle retrieval}) on the WebQA validation set in Table~\ref{tab:webqa_golden}. 
SKURG performs significantly better with oracle retrieval, indicating high retrieval performance benefits QA. 
We hypothesize that the degradation in QA performance on WebQA when using the retrieval module may be due to low retrieval recall, which prevents the answer generation stage from attending to the correct sources. Also note that for w/o RG-Dec, the decoder could still attend to sources and their shared entities from EF-Enc.
Using gold alignment between head entities and sources instead of alignments predicted by EF-Enc (as shown in Row2 of Table~\ref{tab:webqa_golden}) slightly improves the QA performance on WebQA, indicating that more accurate entity alignments could enhance the performance of SKURG.
We also observe that SKURG achieves larger gains on MultimodalQA than on WebQA, probably due to two reasons.
Firstly, the ablation study on MultimodalQA in Table~\ref{tab:multimodal_ablation} shows that SKURG is particularly effective on multi-modal QA. 
WebQA has image and text input sources but only single-modal questions, limiting the ability of SKURG on learning cross-modality alignment and connections hence reducing the benefits of SKURG. 
Note that SKURG is specifically designed for multi-modal QA, and results in Table~\ref{tab:multimodal_ablation} and most results in Table~\ref{tab:webqa_ablation} affirm the significance of both EF-Enc and RG-Dec in enhancing multi-modal QA performance, while on single-modal QA, these components bring smaller gains or cause slight degradations.
Secondly, SKURG is more powerful at generating short key information (Section~\ref{sec:answer_length}), which is more suitable for the short text spans of answers in MultimodalQA compared to the complete-sentence answers in WebQA.

\subsubsection{\textbf{Effect of Answer Length}}
\label{sec:answer_length}


\begin{figure}
  \centering
  \includegraphics[width=0.9\linewidth]{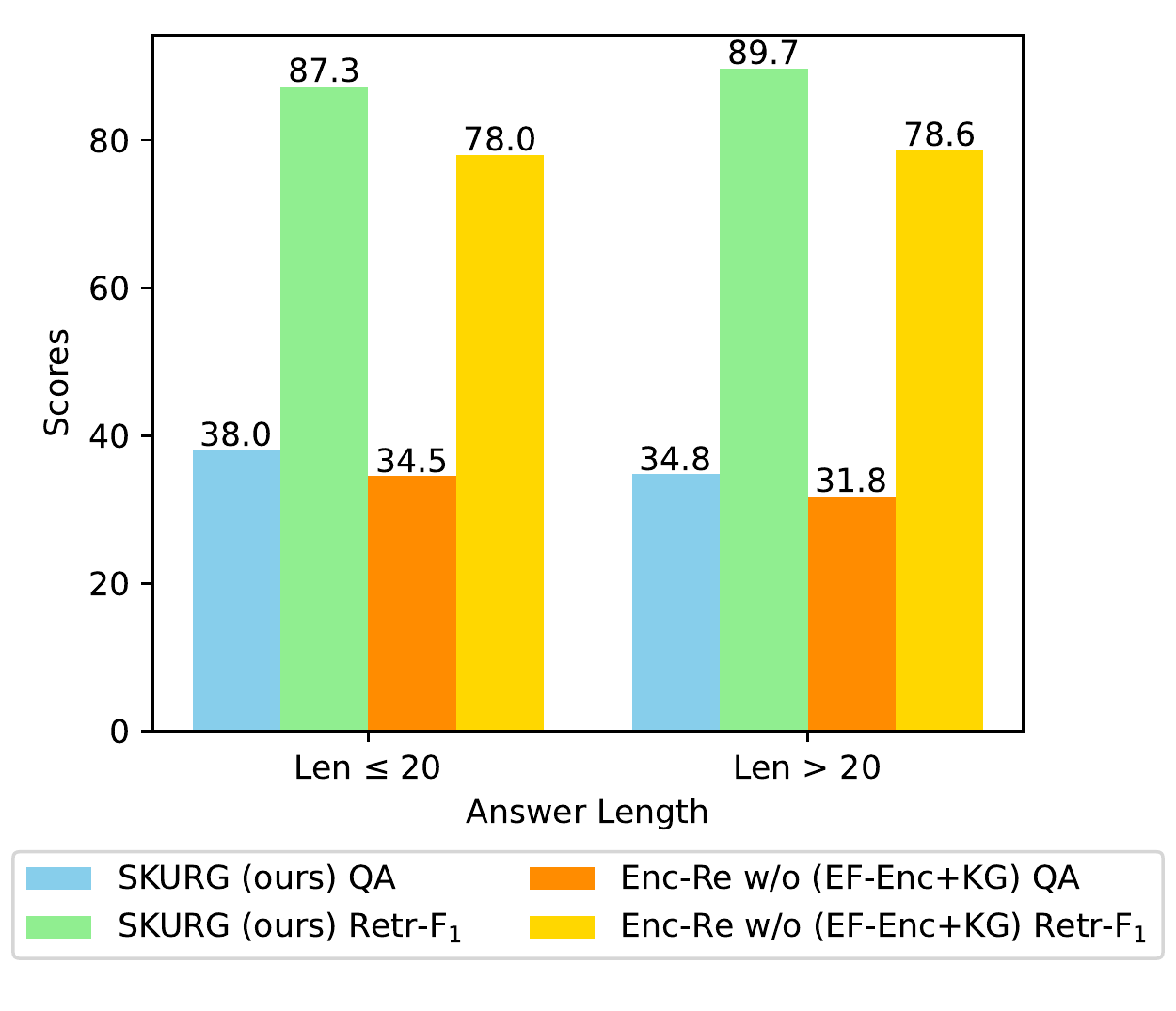}
  \caption{Results of SKURG and baseline (Table~\ref{tab:multimodal_ablation} Row7) on different answer lengths on WebQA \emph{validation set}.}
  \label{fig:answer_length}
\end{figure}

Figure~\ref{fig:answer_length} compares SKURG and the baseline Enc-Re w/o (EF-Enc + KG) (Table~\ref{tab:multimodal_ablation} Row7) on short and long answers on WebQA validation set. 
Answer length denotes the number of tokens in the annotated answer.
Both models exhibit better QA but lower retrieval F$_1$ on shorter answers ($\leq$ 20), but lower QA and better retrieval F$_1$ on longer answers ($>$ 20).
Compared to the baseline, SKURG demonstrates greater gain on retrieval F$_1$ for longer answers versus shorter answers (11.1 vs. 9.3 abs.), but smaller gain in QA score (3.0 vs. 3.5 abs.), suggesting that SKURG is better at generating shorter answers. Using a stronger language model such as T5-base in SKURG might enhance its QA-FL for generating long answers, but it may not necessarily improve the accuracy of key entities in the answers and is not the focus of this paper.
We present the analysis of the trend of better retrieval F1 on longer answers in Appendix~\ref{sec:appendix_ans_len_retrieval}.

\subsubsection{\textbf{More Analyses}}
\label{sec:more_analyses}
We report findings from more analyses in Appendix. Analysis of the effect of the number of sources on SKURG performance is shown in Appendix~\ref{sec:appendix_num_sources}.  The effect of the position of the Entity-Centered Fusion Layer and the distribution of sources with different numbers of entities in the two datasets are analyzed in Appendix~\ref{sec:appendix_ablation_fusion_position} and~\ref{sec:appendix_distribution_of_sources}. 
Appendix~\ref{sec:appendix_ablation_KG}  reports 
the effect of using different knowledge extraction methods for building KG.


\subsection{Case Study}

\begin{figure}
  \centering
  \includegraphics[width=\linewidth]{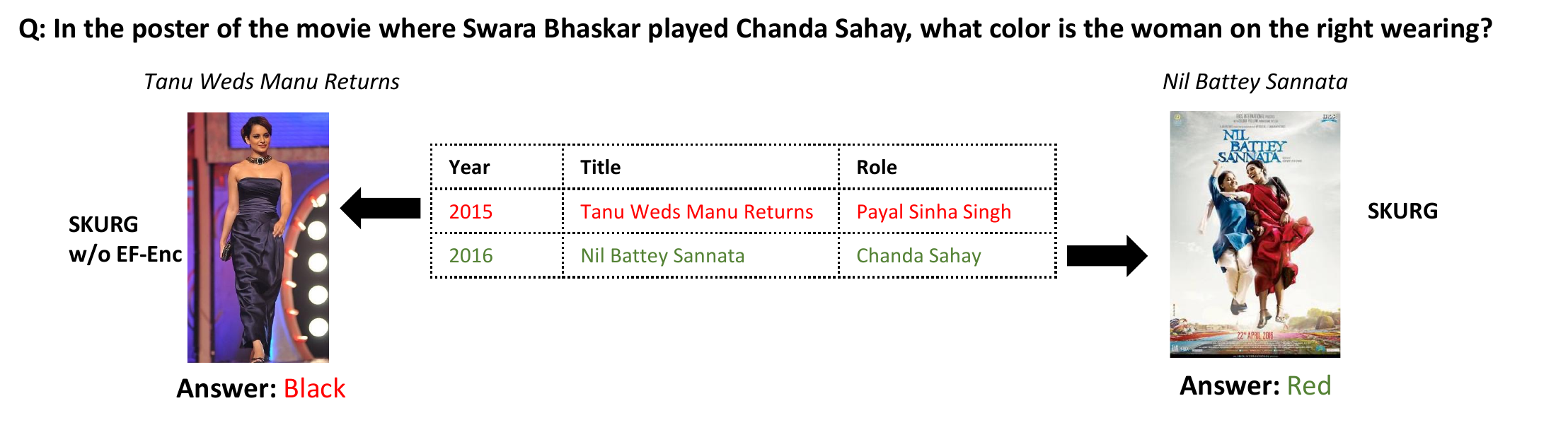}
  \caption{Examples with and without Entity-centered Fusion Encoder.}
  \label{fig:examples_EF_Enc}
\end{figure}

\begin{figure}[tb]
  \centering
  \includegraphics[width=\linewidth]{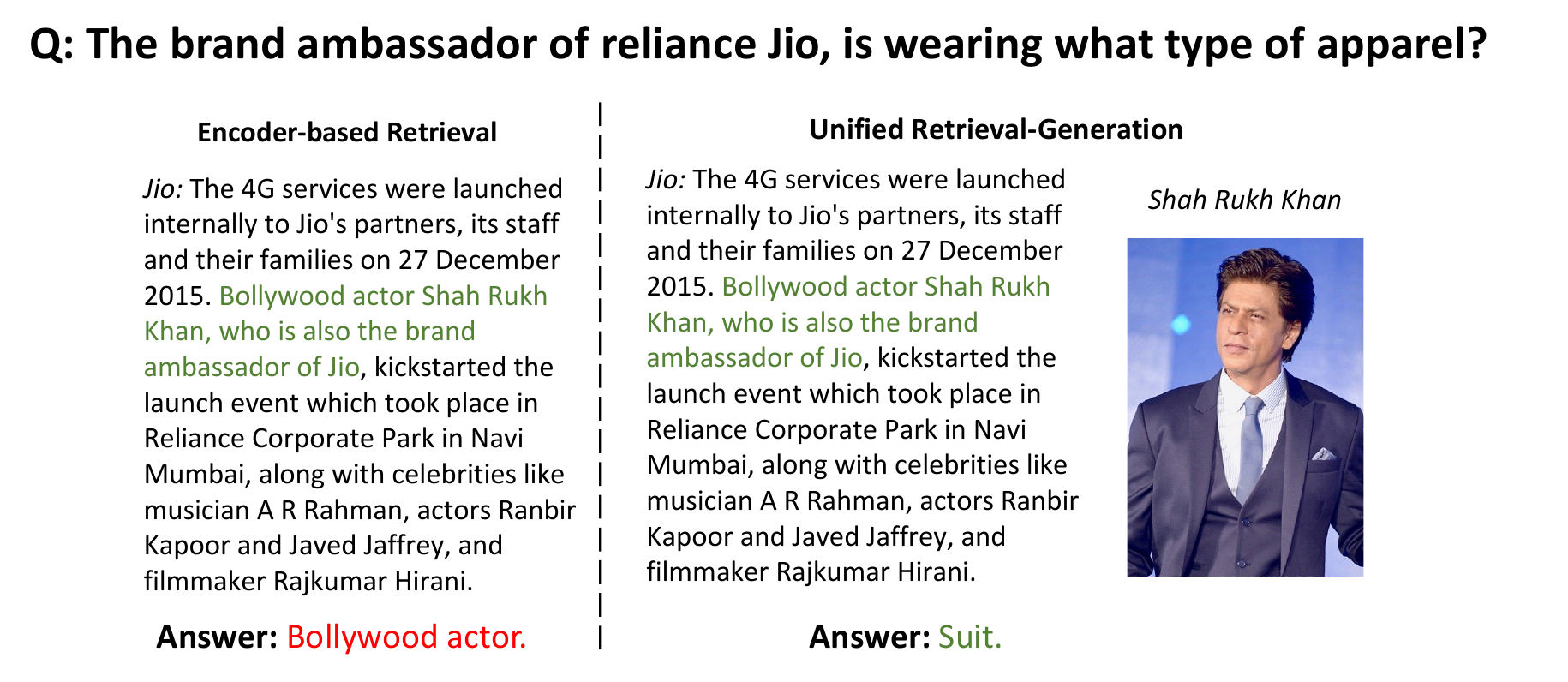}
  \caption{Examples of Encoder-based Retrieval and Unified Retrieval-Generation Decoder.}
  \label{fig:examples_RG_Dec}
\end{figure}

As shown in Figure~\ref{fig:examples_EF_Enc}, despite the table containing the required information for answering the question, without the proposed EF-Enc, the vanilla BART-encoder fails to correctly align the corresponding image with the relevant cell. This misalignment results in retrieving the wrong image and generating an inaccurate answer.
By using EF-Enc to align entities (cells) with input sources, SKURG correctly retrieves the relevant image and generates the correct answer.
Figure~\ref{fig:examples_RG_Dec} compares using Encoder-based Retrieval and our proposed RG-Dec. Both models do not use EF-Enc.
Encoder-based retrieval fails to retrieve the image of \emph{Shah Rukh Khan}, since it is unable to infer that he is the \emph{brand ambassador} of \emph{Reliance Jio} during retrieval.
RG-Dec can model the interdependent relations during retrieval, thus finding the correct evidences and generating the correct answer.

\section{Conclusion AND FUTURE DIRECTION}

We propose a Structured Knowledge and Unified Retrieval-Generation 
model for Multi-modal Multi-hop Question Answering.
By connecting sources through structured knowledge,
our model effectively models interdependent reasoning steps and improves retrieval and answer generation.
The unified retrieval-generation helps integrate intermediate retrieval information for answer generation.
Our model largely outperforms the current SOTA. 
Future work includes improving generation of complete natural-sentence answers and enhancing the capability of handling a large number of sources in open-domain QA by integrating source filtering. 


\begin{acks}
This work is jointly supported by grants: Natural Science Foundation of China (No. 62006061),  Strategic Emerging Industry Development Special Funds of Shenzhen (No. JCYJ20200109113403826).
\end{acks}
  

\bibliographystyle{ACM-Reference-Format}
\balance
\bibliography{reference}

\vfill\eject

\appendix

\section{Loss Function Convergence Analysis}
\label{sec:appendix_loss_function}
We use 5 loss functions during training.
To ensure good convergence, we carefully design the loss functions to align with the objectives of the model. 
Each loss function contributes to the overall training objective, which is to align cross-modal sources, retrieve relevant evidence, and generate accurate answers. This shared objective contributes to the convergence of the loss functions during training.
During training, we use the Adam optimizer with an initial learning rate of $10^{-5}$ and linear decay for the learning rate. We set batch size to 2 and the gradient accumulation steps to 4. This optimization strategy helps the loss functions to converge toward their optimal values.
Furthermore, we monitor and track the values of each loss function throughout the training process. By analyzing these values, we can assess the convergence and effectiveness of each loss function. Based on the best checkpoint on the WebQA validation set, we find each loss value as follows. The alignment loss $\mathcal{L}_{a}$, which determines the alignment between sources and entities in EF-Enc, is lower than 0.8. The confidence loss $\mathcal{L}_{c}$, which estimates the confidence in connecting the source to KG, is lower than 0.5. These values indicate that the alignment process is effectively optimized, as supported by 62.0 F$_1$ for entity alignment in Table~\ref{tab:entity_align_F1_webqa} Row 2.
Regarding the retrieval component, the retrieval loss $\mathcal{L}_{r}$ for RG-Dec, which is responsible for finding the evidence, is lower than 0.2. The loss $\mathcal{L}_{s}$ for learning when to stop retrieval is lower than 0.1. These values demonstrate the successful optimization of the retrieval process, as supported by the evidence retrieval F$_1$ of 88.2 in Table~\ref{webqa_result}.
The answer generation loss $\mathcal{L}_{g}$, based on the negative log-likelihood, is lower than 0.5. The superior answer generation performance of our method compared to other baselines and +1.6 gain on overall QA score over SOTA MuRAG in Table~\ref{webqa_result} demonstrate the effectiveness of $\mathcal{L}_{g}$.
In summary, by carefully designing the loss functions and employing appropriate optimization techniques, we ensure that each loss function achieves good convergence during training.
Monitoring and analysis of loss function values provide insights into their optimization progress and effectiveness in achieving the desired objectives.

\section{Ablation Studies}

\subsection{\textbf{Retrieval Performance with Different Answer Length}}
\label{sec:appendix_ans_len_retrieval}

Table~\ref{tab:answer_length_retrieval} shows the retrieval performance of SKURG and the baseline model Enc-Re w/o (EF-Enc + KG) on the WebQA validation set. For answer lengths greater than 20, the number of input sources also increases (16.88 vs. 16.56).
Both SKURG and the baseline model achieve higher Retrieval F$_1$ scores when the answer length exceeds 20. 
However, the baseline model experiences a \textbf{1.7\%} decrease in recall, while SKURG exhibits an increase in both precision and recall, by \textbf{2.1\%} and \textbf{2.6\%}, respectively.
We hypothesize that with longer answer lengths and more input sources, the content differences between sources become more pronounced. 
The baseline model, which retrieves sources independently, faces challenges in linking information across evidence, resulting in a decrease in average true positive per question (from 1.47 to 1.45) and average false positive per question (from 0.56 to 0.5). 
And thus recall decreases while precision increases.
In contrast, SKURG's ability to build knowledge graphs based on the input sources enables it to create a more comprehensive KG as the number of sources increases. 
As a result, SKURG can effectively model relationships between sources.
This allows SKURG to effectively model relationships between sources, leading to an increase in the average true positives per question (from 1.5 to 1.55) and a decrease in the average false positives per question (from 0.19 to 0.16).
These improvements contribute to enhanced retrieval performance, resulting in improvements across all retrieval metrics.

\subsection{\textbf{Effect of Number of Input Sources}}
\label{sec:appendix_num_sources}

\begin{figure}[b]
  \centering
  \includegraphics[width=\linewidth]{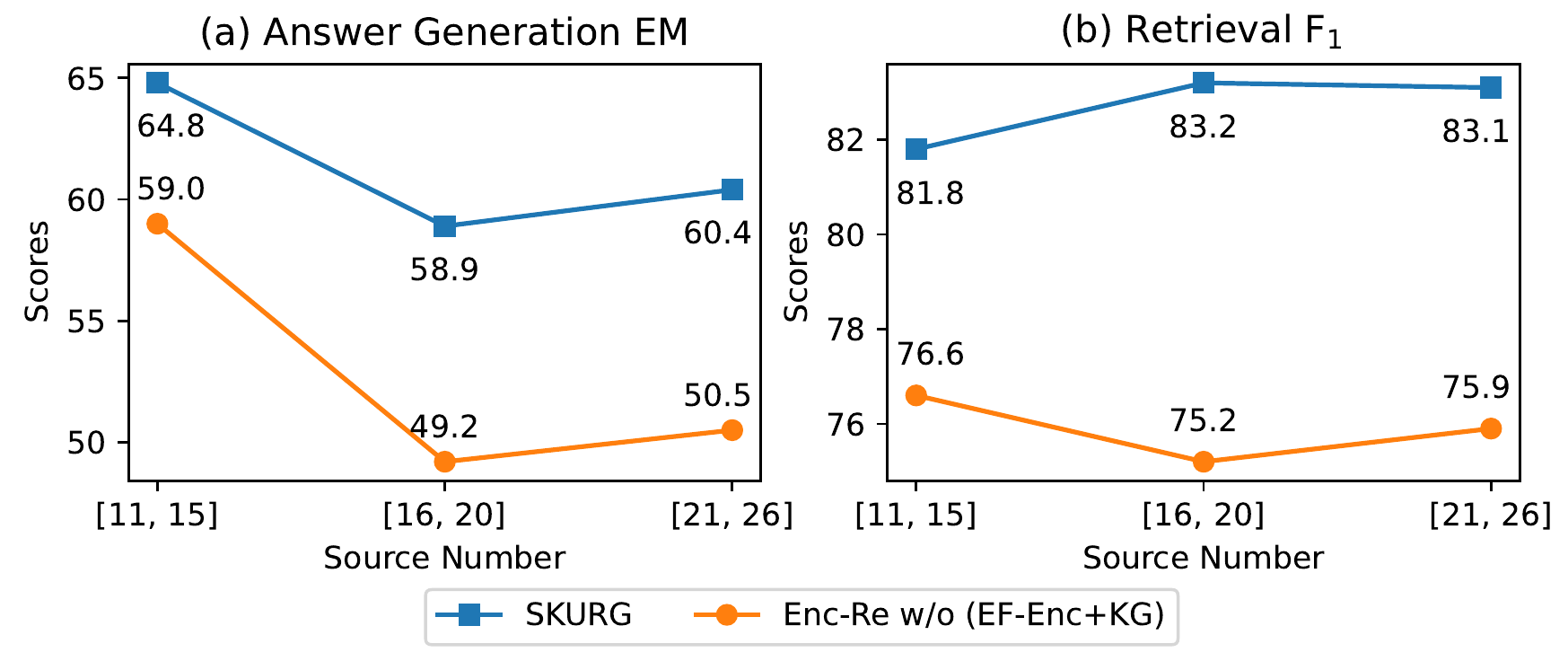}
  \caption{Results of SKURG and baseline (Table~\ref{tab:multimodal_ablation} Row7) against the number of input sources on MultimodalQA \emph{validation set}.}
  \label{fig:num_sources}
\end{figure}

Figure~\ref{fig:num_sources} shows performance of SKURG and the baseline model
Enc-Re w/o (EF-Enc + KG) on MultimodalQA validation set on different numbers of input sources ($n$). Regardless of $n$, SKURG consistently outperforms the baseline on Answer Generation EM and Retrieval F$_1$. SKURG is more effective on larger $n$, with absolute gains of \textbf{5.8}, \textbf{9.7}, \textbf{9.9} on EM score, and \textbf{5.2}, \textbf{8.0}, \textbf{7.2} on Retrieval F$_1$ for $n$ in [11-15], [16-20], [21-26], respectively.
Moreover, Figure~\ref{fig:num_sources} (b) highlights the trend where the retrieval performance of SKURG improves as the number of input sources increases.
This observation aligns with our hypothesis presented in Appendix~\ref{sec:appendix_ans_len_retrieval}  that more input sources contribute to a more complete KG, resulting in enhanced retrieval performance for SKURG.

\begin{table}[b]
\caption{Average input sources per question and retrieval performance of SKURG and baseline (Table~\ref{tab:multimodal_ablation} Row7) on different answer lengths on WebQA \emph{validation} set. Ans-Len refers to the number of tokens in the annotated answer.}
\centering
    \resizebox{\linewidth}{!}{
        \begin{tabular}{cclcccc}
        \toprule
               Ans-Len & Source Num.  &Model   & Precision $\uparrow$       & Recall $\uparrow$   & F$_1\uparrow$           \\ \midrule
            \multirow[m]{2}{*}{Len $\leq$ 20}  &\multirow[m]{2}{*}{16.56}    &SKURG (\textbf{ours})        & 88.5\%	&86.2\%	&87.3\%\\
              &   &Enc-Re w/o (EF-Enc+KG)   & 72.3\%	&84.7\%	&78.0\%\\\midrule
            \multirow[m]{2}{*}{Len $>$ 20}  &\multirow[m]{2}{*}{16.88}    &SKURG \textbf{(ours)}       & 90.6\%   & 88.8\%  & 89.7\%   \\
             &   &Enc-Re w/o (EF-Enc+KG)  & 74.7\% & 83.0\%  & 78.6\%   
                \\\bottomrule
        \end{tabular}
    }
\label{tab:answer_length_retrieval}
\end{table}

\subsection{\textbf{Effect of Position of Entity-Centered Fusion Layer}}
\label{sec:appendix_ablation_fusion_position}

Table~\ref{tab:multimodalqa_index} shows the effect of fusion layer position in the encoder for EF-Enc.
The first line denotes not adding the fusion layer at all.
The encoder contains 6 Transformer layers, and placing the fusion layer in the 4th layer achieves the the best EM and F$_1$ scores.
Placing the fusion layer in the 3rd or 5th layer results in a slight decrease in the QA performance, while placing it in the first or last layer of the encoder leads to a more notable decline. 
We hypothesize that early fusion may hinder learning from the knowledge graph, and late fusion may not fully update the knowledge graph representations.

\begin{table}[t]
\caption{Results on MultimodalQA \emph{validation set} from placing the Fusion Layer at different layers of the BART-encoder for entity-centered fusion encoder in SKURG.}
\centering
   \resizebox{\linewidth}{!}{
        \begin{tabular}{cccccccccc}
        \toprule
         \multirow[m]{2}{*}{Layer}        &\multicolumn{3}{c}{Multi-modal}  &\multicolumn{3}{c}{Single-modal}   &\multicolumn{3}{c}{All}\\
                                  & EM        &F$_1$    &Retr-F$_1$  & EM        &F$_1$   &Retr-F$_1$   & EM   &F$_1$ &Retr-F$_1$ \\ \midrule
               --     & 45.9    & 50.5 & \textbf{80.6}  & \textbf{66.7} &\textbf{70.5}  &86.7  &57.2   &61.3  &\textbf{83.2}   \\      
               1     & 50.6    &54.9 & 79.2  & 65.2     &  69.3  &86.5  &58.5   &62.7  &82.2   \\
               3       &51.9  &  \textbf{57.3}   &80.0   & 65.8 & 69.8   &86.7 &  59.4   &  \textbf{64.0}   &82.8 \\
              4       &\textbf{52.5} &57.2   &\textbf{80.6}  & 66.1 &69.7 &86.7 &\textbf{59.8} &\textbf{64.0}  &\textbf{83.2}   \\
            5        &  51.4    &  56.1      & 80.0 & 65.2    & 69.1  & \textbf{87.1}   &58.8   &63.1  &82.9     \\
              6      &  50.1    &  55.1      & 79.7  & 65.2    & 68.9  & 86.3   &58.3   &62.6  &82.5     \\\bottomrule
        \end{tabular}
    }
\label{tab:multimodalqa_index}
\end{table}

\begin{table}[t]
\caption{Numbers of entities \emph{per source} in MultimodalQA and WebQA datasets.}
\centering
      \resizebox{\linewidth}{!}{
        \begin{tabular}{lccccc}
        \toprule
                Dataset            & Source Num.      & No Ent.    & Single Ent.  & Multiple Ent.	  & Avg. Ent. per Source        \\ \midrule
                  MultimodalQA (Given Table)	&495,320	&47.5\%	&46.5\%	&6.0\%	&0.66 \\
                  MultimodalQA (Extracted KG)	&495,320	&52.4\%	&33.7\%	&13.9\%	&0.66\\
                  WebQA (Extracted KG)	&1,122,200	&22.7\%	&31.0\%	&46.3\%	&1.6  \\\bottomrule
        \end{tabular}
    }
\label{tab:num_of_entities}
\end{table}

\subsection{\textbf{Distribution of Sources with Different Numbers of Entities}}
\label{sec:appendix_distribution_of_sources}

Table~\ref{tab:num_of_entities} shows distributions of sources with different numbers of entities. For the MultimodalQA dataset, both in the given tables and with the extracted KG, the majority of sources have either no entities or a single entity. For the WebQA dataset, with the extracted KG (no tables are given), the distribution is different. A significant portion of sources (46.3\%) contains multiple entities. 
It is important to highlight that SKURG does not utilize multiple entities per source. In cases where a source contains multiple head entities, we randomly select one entity to be aligned. As a result, SKURG currently does not take advantage of multiple entities per source.
We plan to explore strategies to effectively utilize multiple head entities and investigate the influence on the model performance in future work.

\subsection{\textbf{Effect of Knowledge Extraction Methods}}
\label{sec:appendix_ablation_KG}

We investigate the effect of various knowledge extraction methods on the performance of SKURG. 
Table~\ref{tab:entity_align_F1_multimodal} shows the average numbers (per question) of head entities, relations between head entities and tail entities, and head entity-source pairs, and the F$_1$ score of entity alignment of different knowledge extraction methods on MultimodalQA validation set\footnote{Note that Table~\ref{tab:num_of_entities} shows the numbers of entities per source (instead of per question).}.
Table~\ref{tab:multimodalqa_RE} shows the results of different knowledge extraction methods on the MultimodalQA validation set.
\textbf{Given Table Only} refers to using just the tables provided in MultimodalQA for alignment and fusion.
ELMo-based NER~\cite{Peters2017SemisupervisedST} fine-tunes ELMo~\cite{peters-etal-2018-deep} to extract entities.
OpenNRE~\cite{opennre} fine-tunes BERT for relation extraction based on the NER results.
UIE~\cite{lu-etal-2022-unified} uses T5~\cite{t5} to generate entities and relations via structural schema instructor.
Table~\ref{tab:multimodalqa_RE} shows that combining ELMo-based NER with OpenNRE relations achieves the best retrieval performance. 
Using UIE entities and relations slightly decreases F$_1$ and retrieval performance, possibly due to fewer relations and head entity-source provided by UIE and the lower F$_1$ score of entity alignment.
Table~\ref{tab:entity_align_F1_webqa} and Table~\ref{tab:webqa_RE} show extraction details and results of knowledge extraction methods on WebQA.
Wiki means matching extracted entities with Wikidata~\cite{vrandevcic2014wikidata} to obtain relations. 
Note that we do not use Wikidata on MultimodalQA because it is annotated on Wikidata.
Results in Table~\ref{tab:webqa_RE} indicate that using Wikidata for relation extraction yields the best QA performance, indicating that accurate relations can help improve the effectiveness.

\begin{table}[th]
\caption{Average numbers (per-question) of head entities, relations between head entities and tail entities, head entity-source pairs and F1 score of entity alignment on MultimodalQA \emph{validation set} using different knowledge extraction methods.}
\centering
    \resizebox{\linewidth}{!}{
        \begin{tabular}{lcccc}
        \toprule
                Extraction Method             & Ent.   & Relation  & Ent. Source Pair    & Align-F$_1$        \\ \midrule
                Given Table Only    & 14.0   &  63.1    &10.0    & 85.4 \\
                ELMo-based NER    & 8.5 &  --    &9.6  &72.3  \\
                \textbf{ELMo-based NER + OpenNRE}      & 8.4   &  55.5     &10.0   & 71.1 \\
                UIE    &  8.8   &  28.9  &7.1    &  49.8 \\\bottomrule
        \end{tabular}
    }
\label{tab:entity_align_F1_multimodal}
\end{table}

\begin{table}[thp]
\caption{Results for using different relation extraction methods in SKURG on MultimodalQA \emph{validation} set.}
\centering
   \resizebox{\linewidth}{!}{
        \begin{tabular}{lccccccccc}
        \toprule
         \multirow[m]{2}{*}{Extraction Method }     &\multicolumn{3}{c}{Multi-modal}    &\multicolumn{3}{c}{Single-modal}    &\multicolumn{3}{c}{All}\\
                                  & EM        &F$_1$    &Retr-F$_1$  & EM        &F$_1$   &Retr-F$_1$   & EM   &F$_1$ &Retr-F$_1$ \\ \midrule
                Given Table Only    & 51.6    &55.6 &79.8   & 65.8     &  \textbf{70.3}  &\textbf{87.1}  &59.3   &63.5  &82.9   \\
                + ELMo-based NER      &\textbf{52.8}  &  57.1   &79.8   & \textbf{66.3} & 69.5   &86.7  &  \textbf{60.1}   &  63.8   &82.8  \\
                \textbf{+ ELMo-based NER + OpenNRE}     &52.5 &\textbf{57.2}   &\textbf{80.6}   & 66.1 &69.7 &86.7 &59.8 &\textbf{64.0}  &\textbf{83.2}   \\
                + UIE       &  51.0    &  55.2      & 78.6  & 67.1    & 70.1  & 86.4   &59.7   &63.2  &82.0     \\\bottomrule
        \end{tabular}
    }
\label{tab:multimodalqa_RE}
\end{table}

\begin{table}[b]
\caption{Average numbers (per-question) of head entities, relations between head entities and tail entities, and head entity-source pairs and F1 score of entity alignment on WebQA \emph{validation set} using different knowledge extraction methods.}
\centering
    \resizebox{\linewidth}{!}{
        \begin{tabular}{lcccc}
        \toprule
                Extraction Method             & Ent.   & Relation  & Ent. Source Pair    & Align-F$_1$        \\ \midrule
                ELMo-based NER    & 13.6    &  --      & 23.1    &58.0 \\
                \textbf{ELMo-based NER + OpenNRE}       & 10.6    &  76.9      & 18.5   &62.0   \\
                ELMo-based NER + Wiki      & 11.4  & 11.4      & 22.0  & 58.3 \\
                UIE    & 13.2     & 51.9      & 15.9   &51.6  \\\bottomrule
        \end{tabular}
    }
\label{tab:entity_align_F1_webqa}
\end{table}

\begin{table}[b]
\caption{Results for using different relation extraction methods in SKURG on WebQA \emph{validation} set.}
\centering
    \resizebox{\linewidth}{!}{
        \begin{tabular}{lcccc}
        \toprule
                Extraction Method            & QA-FL       & QA-Acc     & QA  & Retr-F$_1$           \\ \midrule
                  ELMo-based NER   & 47.5    &  63.4      &37.3    & 87.2 \\
                  \textbf{ELMo-based NER + OpenNRE}      &46.8 &\textbf{64.3}  &37.1  &\textbf{87.5}\\
                  ELMo-based NER +  WiKi  &\textbf{47.6}  &63.3   &\textbf{37.8}  &87.1    \\
                  UIE &46.4 &62.5  &37.5 &86.0  \\\bottomrule
        \end{tabular}
    }
\label{tab:webqa_RE}
\end{table}

\end{document}